\documentclass[runningheads]{llncs}

\usepackage{eccv}

\usepackage{eccvabbrv}

\usepackage{graphicx}
\usepackage{booktabs}
\usepackage{multirow}
\usepackage[accsupp]{axessibility}  
\usepackage{xcolor} 
\usepackage[table,xcdraw]{xcolor}
\usepackage{makecell}

\usepackage{hyperref}

\usepackage{orcidlink}

\begin{document}

\title{Progressive Representation Learning for Multimodal Sentiment Analysis with Incomplete Modalities} 

\titlerunning{Abbreviated paper title}

\author{Jindi Bao\inst{1} \and
Jianjun Qian\inst{1} \and
Mengkai Yan\inst{2}\and
Jian Yang\inst{1}
}
\authorrunning{J.Bao et al.}

\institute{Nanjing University of Science and Technology, Nanjing 210094, China \and
Hohai University, Nanjing 211100, China
}

\maketitle

\begin{abstract}
Multimodal Sentiment Analysis (MSA) seeks to infer human emotions by integrating textual, acoustic, and visual cues.
However, existing approaches often rely on all modalities are completeness, whereas real-world applications frequently encounter noise, hardware failures, or privacy restrictions that result in missing modalities.
There exists a significant feature misalignment between incomplete and complete modalities, and directly fusing them may even distort the well-learned representations of the intact modalities.
To this end, we propose PRLF, a Progressive Representation Learning Framework designed for MSA under uncertain missing-modality conditions.
PRLF introduces an Adaptive Modality Reliability Estimator (AMRE), which dynamically quantifies the reliability of each modality using recognition confidence and Fisher information to determine the dominant modality. 
In addition, the Progressive Interaction (ProgInteract) module iteratively aligns the other modalities with the dominant one, thereby enhancing cross-modal consistency while suppressing noise.
Extensive experiments on CMU-MOSI, CMU-MOSEI, and SIMS verify that PRLF outperforms state-of-the-art methods across both inter- and intra-modality missing scenarios, demonstrating its robustness and generalization capability.
\end{abstract}

\section{Introduction}
\label{sec:intro}
Multimodal Sentiment Analysis (MSA) has emerged as a crucial research topic in affective computing, aiming to infer human emotions by integrating heterogeneous cues from text, speech, and visual signals \cite{r2,r3,r4,r5,r6,r7,r58,r59}.
By leveraging the complementarity among different modalities, MSA has achieved remarkable progress with the development of deep learning \cite{r8,r9}.
However, most existing approaches rely on an ideal assumption that all modalities are consistently available in both the training and inference stages \cite{r10}.
In practical scenarios, this assumption rarely holds true, as various factors such as environmental noise, hardware malfunction, data transmission failure, or privacy restrictions may cause uncertain and incomplete modality inputs \cite{r11}.
Such imperfect multimodal data often lead to degraded feature representations and unstable emotional understanding, posing a critical challenge for building robust MSA systems \cite{r12}.
\begin{figure}
  \centering
  \begin{subfigure}{0.48\linewidth}
   \includegraphics[width=\linewidth]{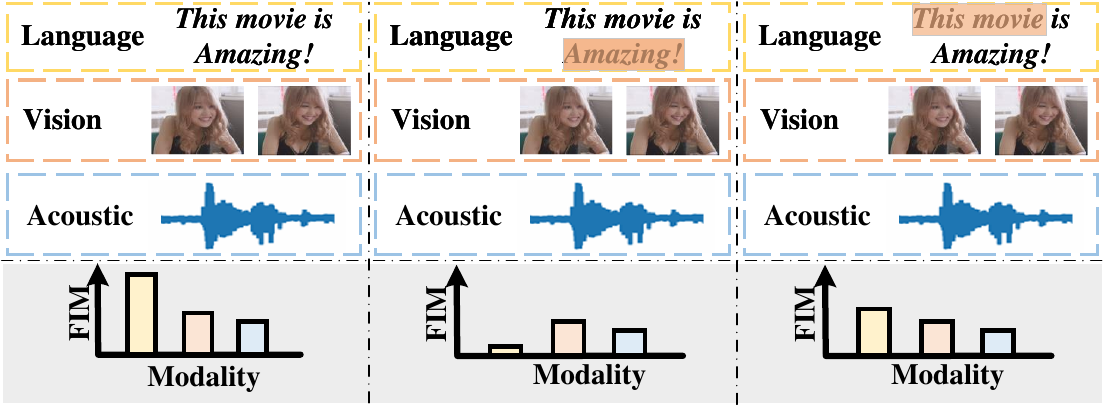}
    \caption{Variation of Modality Feature Information under Different Missing-Data Conditions.}
    \label{fig1:short-a}
  \end{subfigure}
  \hfill
  \begin{subfigure}{0.48\linewidth}
  \includegraphics[width=\linewidth]{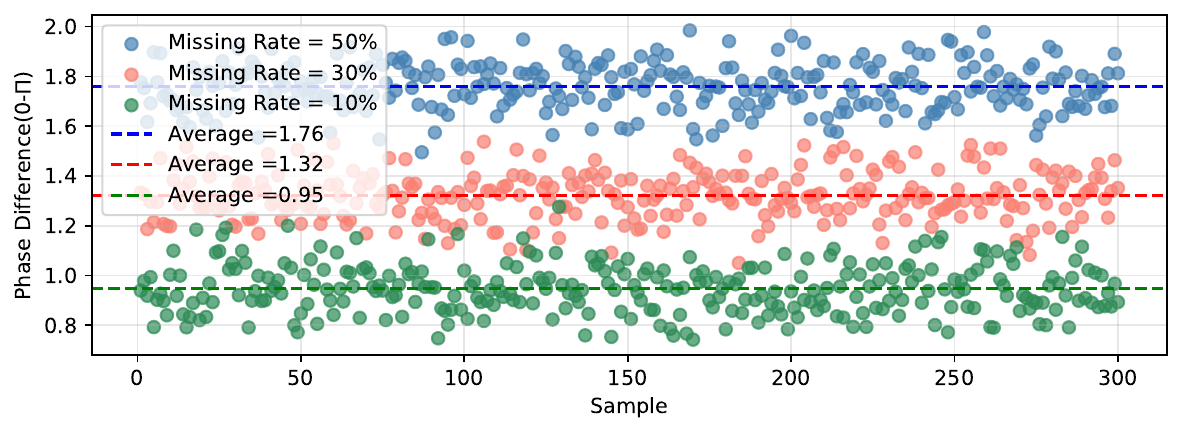}
    \caption{Feature Phase Differences under Different Missing Rates.}
    \label{fig1:short-b}
  \end{subfigure}
  \vspace{-0.2em}
  \caption{Feature information loss and feature phase shift caused by missing modalities.}
  \label{fig:short}
  \vspace{-1.7em}
\end{figure}

Recently, growing attention has been directed toward addressing the challenge of incomplete modalities in MSA, and existing research can generally be categorized into two main directions: (\textbf{i}) Generative approaches, which recover the missing modalities by leveraging the information from the available ones \cite{r13,r14,r15,r16,r21}; (\textbf{ii}) Distillation-based methods, which transfer knowledge from complete-modality models to the models under missing-modality conditions \cite{r17,r18,r19,r20,r22}.
Although two strategies have made progress, these methods still follow fusion paradigms similar to traditional approaches and fail to account for the importance differences between missing and complete modalities.
We use the Fisher Information Matrix (FIM) \cite{r1,r35} to evaluate the impact of missing modalities.
As shown in Fig.~\ref{fig1:short-a}, the loss of critical data in a specific modality can lead to a significant reduction in information content, whereas the absence of non-critical data has a relatively minor impact. 
Therefore, identifying the importance differences among modalities under various missing-data conditions is crucial for optimizing model design and fusion strategies.

Meanwhile, the impact of missing data is often difficult to eliminate, and existing methods generally overlook its effect during multimodal fusion.
To investigate the impact of modality missing rates on the extracted features, we selected 300 samples and calculated the phase differences between the features obtained under various missing rates and those extracted from the complete modality, as shown in Fig.~\ref{fig1:short-b}.
Here, phase difference refers to the angular deviation in the high dimensional feature space.
We observed that as the missing rate increases, the phase difference between the extracted features and those obtained from the complete modality also increases.
Therefore, the impact of missing data during modality interaction and fusion is manifested as phase shifts in the high-dimensional feature space.

Based on the above discussion, we propose a \textbf{P}rogressive \textbf{R}epresentation \textbf{L}earning \textbf{F}ramework (PRLF) for multimodal sentiment analysis under uncertain missing-modality conditions.
The framework enables the model to adaptively learn the importance of each modality for every sample under various missing-data conditions, identify the dominant and auxiliary modalities, and fuse them in a progressive manner.
In particular, we design an Adaptive Modality Reliability Estimator (AMRE), where a router network dynamically evaluates modality reliability based on recognition confidence and Fisher information, and subsequently leverages the dominant modality to guide cross-modal interaction and fusion.
Further considering that missing data disrupts the feature distribution in high-dimensional space, direct cross-modal fusion may distort the representations of the intact modalities.
To this end, we design a Progressive Interaction Module (ProgInteract), which performs iterative interactions to gradually align the feature distribution of auxiliary modalities with that of the dominant modality, thereby enhancing emotion information extraction and mitigating noise interference.
Specifically, at the early stage of training, the model encounters substantial noise and therefore focuses on extracting intra-modal features while performing only limited cross-modal interactions.
As the iteration progresses and the intra-modal representations become more stable, strengthening cross-modal interactions encourages auxiliary modalities to align with the dominant one, ultimately enabling effective multimodal fusion.

Our main contributions are summarized as follows:
\vspace{-0.2em}
\begin{itemize}
    \item We propose a Progressive Interaction Module (ProgInteract) that iteratively aligns auxiliary modality features with the dominant modality, enabling noise-robust and adaptive cross-modal fusion under missing-data conditions.
    \item We propose Adaptive Modality Reliability Estimator to evaluate the effectiveness of each modality and adaptively determines the dominant modality.   
    \item We evaluate our method on standard multimodal datasets, including CMU-MOSI, CMU-MOSEI and SIMS, and the results demonstrate that our approach outperforms or matches the state-of-the-art.
\end{itemize}

\section{Related Work}

\subsection{Multimodal Sentiment Analysis}
Multimodal Sentiment Analysis (MSA) is a task designed to perceive and process heterogeneous data such as language, audio, and visual information, with the goal of understanding and interpreting human emotional states\cite{r31,r32,r33}.
Current mainstream research primarily focuses on developing more effective modality fusion strategies and interaction mechanisms to further enhance the overall performance of MSA \cite{r26,r27,r28,r29,r30}.
For instance, MAMSA \cite{r23} adjusts the contribution of text and image features through adaptive attention and hierarchical fusion to effectively extract and integrate sentiment-related information across modalities.
DB-MPCA \cite{r24} integrates raw acoustic and visual data into a language model via multimodal pretraining and cross-modal attention, preserving textual dominance for more balanced fusion.
DLF \cite{r25} disentangles shared and modality-specific features, refines them with geometric constraints, and enhances language representations via language-guided cross-attention for improved sentiment analysis.
However, these methods assume the availability of all modalities during inference, making them difficult to apply in real-world scenarios where data from certain modalities may be missing, incomplete, or corrupted.

\subsection{Multimodal Sentiment Analysis with Incomplete Data}
In multimodal sentiment analysis, the missing modality problem is typically tackled using two main strategies: (1) Generative approaches and (2) Distillation-based approaches. 
Generative approaches focus on reconstructing absent features and semantic information by modeling the distributions of the observed modalities.
For example, DiCMoR \cite{r34} using class-specific normalizing flows to align distributions between recovered and true data.
MRAN \cite{r13} aligns and reconstructs modalities to enhance robustness when the text modality is missing.
Distillation-based approaches leverage knowledge from models trained on complete modalities to steer the training of models that must operate with incomplete or missing modalities.
For example, UMDF \cite{r18} transfers cross-modal knowledge through multi-grained interaction and dynamic feature integration for robust sentiment analysis under uncertain missing modalities.
CorrKD \cite{r17} leverages contrastive, prototype-based, and consistency distillation to reconstruct missing semantics and enhance robustness under uncertain missing modalities.
However, existing methods fail to account for the directional consistency between features from missing modalities and those from complete modalities, making it difficult to suppress noise interference during multimodal interaction and fusion.
\vspace{-0.5em}

\section{Method}
\vspace{-0.5em}
\subsection{Problem Formulation}
\vspace{-0.3em}
For a multimodal video segment represented as $\textbf{\textit{S}}=[\textbf{\textit{X}}_V,\textbf{\textit{X}}_A,\textbf{\textit{X}}_L]$, the three components $\textbf{\textit{X}}_V \in \mathbb{R} ^{{T_V} \times {d_V}}$, $\textbf{\textit{X}}_A \in \mathbb{R} ^{{T_A} \times {d_A}}$, $\textbf{\textit{X}}_L \in \mathbb{R} ^{{T_L} \times {d_L}}$ correspond to the visual, acoustic, and language modalities, typical of multimodal data, respectively.
Here, $T_m(\cdot)$ and $d_m(\cdot)$ denote the sequence length and feature dimension of modality $m \in \{\textbf{\textit{V}},\textbf{\textit{A}},\textbf{\textit{L}}\}$.
To simulate realistic multimodal scenarios, we consider two types of missingness: (i) intra-modality missingness, referring to the absence of certain frame-level features within a modality sequence; and (ii) inter-modality missingness, referring to the complete absence of modalities.

\subsection{Overall Framework}
Fig. ~\ref{fig1:short-a} illustrates the workflow of PRLF.
PRLF consists of two key components: the Adaptive Modality Reliability Estimator (AMRE) and the Progressive Interaction module (ProgInteract).
The AMRE module identifies the dominant modality based on classification confidence and Fisher information.
The ProgInteract module gradually aligns the feature distributions of the auxiliary modalities with that of the dominant modality through iterative interactions.
The PRLF model takes three modalities as input: visual (V), acoustic (A), and language (L).
Before entering the AMRE module, a dedicated encoder $f_m=\varepsilon _m(\textbf{\textit{X}}_m)$ is employed for each modality to extract its feature representation. 
The resulting $f_m$ serves as the unimodal feature, which is used both as input to the AMRE module and for the subsequent ProgInteract to enable collaborative learning and feature alignment across modalities.

\begin{figure*}
\vspace{-1.0em}
  \centering
   \includegraphics[width=0.95\linewidth]{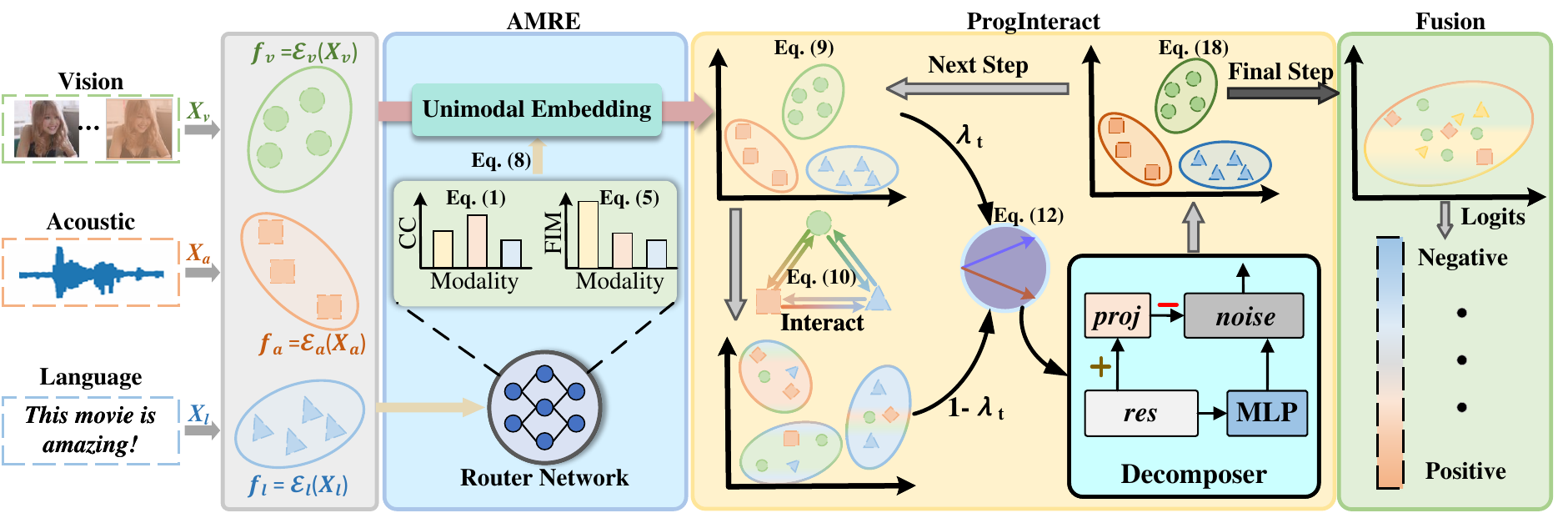}
    \vspace{-0.2em}
    \caption{The structure of PRLF, which consists of two key components: Adaptive Modality Reliability Estimator (AMRE) and Progressive Interaction module (ProInteract).}
    \label{fig1:short-a}
     \vspace{-1.6em}
\end{figure*}

\subsection{Adaptive Modality Reliability Estimator}
\textbf{Confidence-based Modality Importance (CMI).}
To dynamically assess the importance of each modality for different samples, many multimodal methods compute modality importance weights based on the recognition accuracy and confidence of each unimodal model on the given sample \cite{r36,r37,r38}.
Based on this, we design a routing network that effectively assigns an individually independent classification head to each modality.
For a given sample, the classification confidence of modality $m$ for the correct class is obtained as $\alpha^{(i)} _m = \hbar_m (f^{(i)}_m)$, where $m \in \{\textbf{\textit{V}},\textbf{\textit{A}},\textbf{\textit{L}}\}$, $\hbar_m$ is the classification head, $i$ represents sample, and $\alpha$ denotes the confidence for the correct class.
The classification confidences of all modalities are then concatenated and normalized for unified representation, as follows:
\begin{equation}
\alpha^{(i)} = [\alpha^{(i)}_v,\alpha^{(i)}_a,\alpha^{(i)}_l], 
\quad
\hat{\alpha}^{(i)} = \frac{\alpha^{(i)}}{\left\| \alpha^{(i)} \right\|_1}
\end{equation}

\begin{figure}
  \centering
  \begin{subfigure}{0.255\linewidth}
   \includegraphics[width=\linewidth]{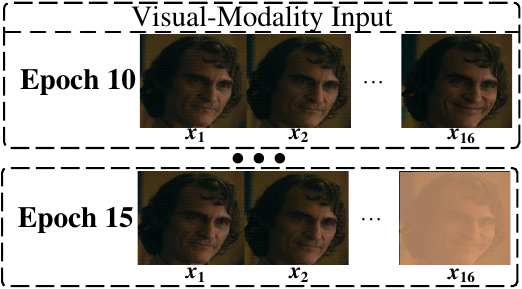}
    \caption{Variation of visual input completeness across training epochs.}
    \label{fig3:short-a}
  \end{subfigure}
  \begin{subfigure}{0.255\linewidth}
  \includegraphics[width=\linewidth]{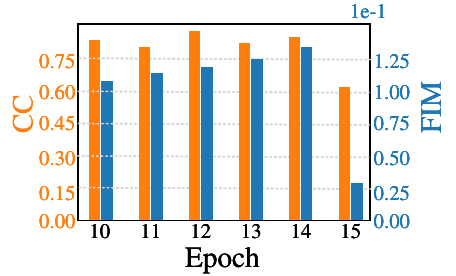}
    \caption{Confidence and Fisher information under missing key frames.}
    \label{fig3:short-b}
  \end{subfigure}
  \begin{subfigure}{0.47\linewidth}
  \includegraphics[width=\linewidth]{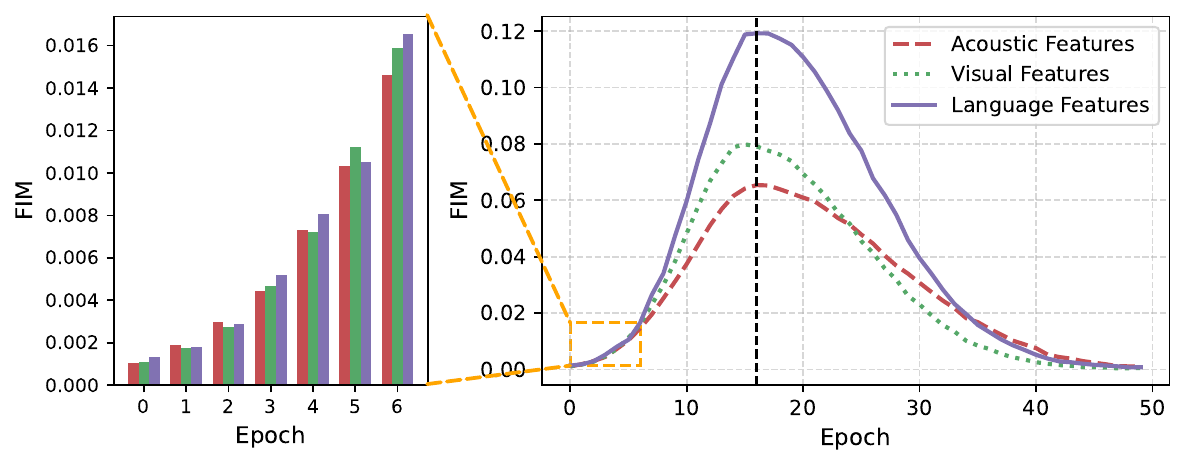}
    \caption{Evolution of Fisher information during training under complete-data inputs.}
    \label{fig3:short-c}
  \end{subfigure}
  \vspace{-1.3em}
  \caption{Analysis of confidence and Fisher information variations caused by missing key frames in the visual modality.}
  \label{fig:short}
  \vspace{-1.3em}
\end{figure}

\noindent
Each classification head is trained with a cross-entropy loss:
\begin{equation}
    \mathcal{L}_{uni} = - \frac{1}{N} \sum_{i=1}^{N} \sum_{m} y^{(i)} \log \sigma(\hbar_m(f^{(i)}_m)),
\end{equation}

\noindent
where $y^{(i)}$ is the ground-truth label and $\sigma(\cdot)$ denotes the softmax function.
However, we found that under missing data conditions, relying solely on classification confidence is insufficient for reliable evaluation.
As shown in Fig.~\ref{fig3:short-a}, in the visual modality, the model receives complete input data during epochs 10–14, while key frames are missing at epoch 15.
Fig.~\ref{fig3:short-b} illustrates the variations in classification confidence and Fisher information. 
During epochs 10–14, the model produces correct predictions with high confidence. 
Yet, even when key frames are missing in epoch 15, it still reports a high confidence score. 
This phenomenon may result from the model’s memorization of facial features \cite{r41,r43}. 
Therefore, relying solely on confidence to identify the dominant modality is unreliable.
Notably, the trace of the Fisher Information Matrix reveals a substantial decline in the effective information contained in the input. 

\textbf{Fisher Information-based Modality Importance (FIMI).}
The trace of the Fisher Information Matrix (FIM) measures the overall sensitivity of model parameters to the output distribution.
It reflects how much effective information the model extracts from the data.
In multimodal learning, for a unimodal encoder $\varepsilon_m{(\cdot)}$ and its corresponding classification head $\hbar_m{(\cdot)}$, the gradient of the model parameters $\theta$ can be expressed as:
\vspace{-0.5em}
\begin{equation}
    g_m = \nabla_\theta \big(\hbar_m \circ \varepsilon_m(\textbf{\textit{X}}_m)\big)
\end{equation}
Based on this, the trace of the FIM for modality $m$ can be written as: 
\vspace{-0.3em}
\begin{equation}
    \mathrm{Tr}(F_m^{(i)}) = \mathrm{Tr}\!\Big(\mathbb{E}_i[g_m^{(i)} {g_m^{(i)}}^\top]\Big) = \mathbb{E}_i\!\big[\| g_m^{(i)} \|_2^2 \big]
\end{equation}

\noindent
where $\mathrm{Tr}$ denotes the trace of a matrix, $\mathbb{E}$ represents the mathematical expectation, $g_m g_m^\top$ denotes the outer product of the gradient vector, and $\| g_m \|_2^2$ represents the squared $\ell_2$ norm of the gradient.
A larger Tr(FIM) indicates that the model’s output is more sensitive to input perturbations, thereby reflecting the amount of effective information contained in that modality.
As shown in Fig. ~\ref{fig3:short-b}, we observe that when key frames are missing, the trace of the Fisher Information Matrix, $\mathrm{Tr}(F_m)$, exhibits a significant decrease. To explain this phenomenon, we provide the following theoretical interpretation.

Assume that the visual modality consists of 16 frames $\mathbf{X}_m=\{x_1,\dots,x_{16}\}$. 
The gradient of the model with respect to the parameters can be approximately decomposed as the sum of contributions from individual frames:
\vspace{-0.7em}
\begin{equation}
\nabla_\theta \log p_\theta(y \mid \mathbf{X}m)
\approx
\sum_{t=1}^{16} g_t,
\end{equation}
\noindent
where $g_t$ denotes the gradient contribution of the $t$-th frame.
Under partial-missing conditions, we introduce a binary mask $\delta_t \in \{0,1\}$ to indicate whether a frame is present. The gradient then becomes:
\vspace{-0.5em}
\begin{equation}
\nabla_\theta \log p_\theta(y \mid \mathbf{X}m)
\approx
\sum_{t=1}^{16} \delta_t g_t.
\end{equation}
\vspace{-0.2em}
Substituting this into the definition of the trace of the Fisher information yields:
\begin{equation}
\mathrm{Tr}(F_m)=\mathbb{E}\!\left[
\left\|
 {\textstyle \sum_{t=1}^{16}}  \delta_t g_t
\right\|_2^2
\right]
\end{equation}

If certain frames contain key semantic or emotional cues, their gradient magnitudes are typically much larger,
\vspace{-0.3em}
\begin{equation}
|g_t^{\text{key}}|_2^2
\gg
|g_t^{\text{non}}|_2^2.
\end{equation}
when key frames are removed ($\delta_t=0$), the dominant contributors to the gradient energy vanish, leading to a substantial reduction in the overall gradient norm and consequently a clear decline in $\mathrm{Tr}(F_m)$. In contrast, removing non-key frames has a relatively minor impact on the Fisher information.

Therefore, we compute the trace of FIM for each modality and form a vector $\beta^{(i)}$ for the $i$-th sample:
{\vspace{-8.95pt}
\begin{equation}
\beta^{(i)} = [\mathrm{Tr}(F_v^{(i)}), \mathrm{Tr}(F_a^{(i)}), \mathrm{Tr}(F_l^{(i)})], 
\quad
\hat{\beta}^{(i)} = \frac{\beta^{(i)}}{\left\| \beta^{(i)} \right\|_1}
\end{equation}
\noindent
The resulting $\hat{\beta }^{(i)}$ is used as the Fisher information-based modality importance for the $i$-th sample.
However, in the early stages of training, the model exhibits weak gradient responses, resulting in low Fisher information across all modalities.
Consequently, it is difficult to provide a reliable estimation of modality importance at this stage, as shown in Fig.~\ref{fig3:short-c}
At this stage, the model’s classification confidence can reflect the relative contribution of each modality.
Although the confidence is not yet stable, a modality with relatively higher confidence tends to capture more discriminative patterns in its feature space.
Therefore, classification confidence can serve as a complementary and indicative cue for assessing modality importance during the early phase of training.


\textbf{Modality Importance Fusion Mechanism.}
Based on the above analysis, we design a modality importance fusion mechanism that allows the classification confidence and Fisher information to complement each other, thereby enabling a more accurate estimation of modality importance.
Inspired by \cite{r42}, we first assess whether FIM can provide a reliable measure of modality importance by evaluating its growth between consecutive training epochs, as follows:
\begin{equation}
    \Delta^{(t,i)}_m =  \frac{\mathrm{Tr}(F_m^{(t,i)}) - \mathrm{Tr}(F_m^{(t-1,i)})}{\mathrm{Tr}(F_m^{(t,i)})}
\end{equation}
where $t$ is the training epoch, $m \in \{\textbf{\textit{V}},\textbf{\textit{A}},\textbf{\textit{L}}\}$.
This ratio measures the increase of the Fisher information in the current epoch relative to the previous epoch.
Dynamically compute the fusion weight between classification confidence and Fisher information based on the relative change in Fisher information $\Delta ^{(t,i)}_m$:
\begin{equation}
    w^{(t,i)} = \sigma \big( \, [\Delta^{(t,i)}_v,\Delta^{(t,i)}_a,\Delta^{(t,i)}_l] \big)
\end{equation}
where $w^{(t,i)}$ represents the fusion weight for the $i$-th sample at training epoch $t$, $\sigma$ denotes the Sigmoid function.
The modality importance derived from classification confidence and that derived from Fisher information are fused according to the dynamically computed weights $w_m^{(t)}$:
\begin{equation}
    \mu^{(t,i)} = (1 - w^{(t,i)}) \, \hat{\alpha}^{(i)} + w^{(t,i)} \, \hat{\beta}^{(i)}
\end{equation}
Accordingly, in early training, the low fusion weight makes the fusion rely more on classification confidence, whereas when Fisher information rises significantly, a higher weight shifts the fusion toward Fisher information.
\subsection{Progressive Interaction Module}
In Fig.~\ref{fig1:short-b}, we observe that the impact of missing data can cause deviations in feature directions within the high-dimensional feature space. 
During the interaction and fusion process, such deviations can negatively affect the final fused features. 
To mitigate this, we propose a progressive interaction module, which does not perform direct feature fusion. 
Instead, it iteratively refines multimodal representations, focusing on unimodal feature enhancement in the early iterations and emphasizing cross-modal interactions in the later ones.

Specifically, in each iteration, the three modal features $f_m$ first perform self-refinement to extract their internal discriminative information:
\begin{equation}
f_m^\text{self} = f_m + \text{Dropout}\Big(\text{ReLU}(f_m W_1 + b_1) W_2 + b_2 \Big)
\end{equation}
Next, cross-modal interactions are performed across all modalities, with the modality importance $\mu_m$ incorporated during the interaction process:
\begin{equation}
f_{m\to n} = \mathrm{softmax}\!\left(
\frac{(\mu_m f_m)(\mu_n f_n)^{\mathbf T}}{\sqrt d}
\right)(\mu_m f_m),
\quad n \in \{n_1,n_2\}
\end{equation}
where $m,n_1,n_2 \in \{\textbf{\textit{V}},\textbf{\textit{A}},\textbf{\textit{L}}\}$, $n_1 \neq n_2 \neq m$, $d$ means the dimension of $f_m$.
We concatenate the features $f_c=[f_{m\to n_1},f_{m\to n_2}]$, and extract the fused features:
\begin{equation}
    f_m^{\text{cross}} = \mathrm{softmax} (\frac{f_cf_c^\mathbf{T} }{\sqrt{d} })f_c
\end{equation}
For unimodal features and cross-modal features, We introduce a time-dependent weighting coefficient $\lambda_t$ to balance the contribution between the unimodal feature $f_m^{self}$ and the cross-modal feature $f_m^{cross}$:
\begin{equation}
    f_m^{\text{fuse},t} = \lambda_t f_m^\text{self} + (1-\lambda_t) f_m^{\text{cross}}
\end{equation}
where, $\lambda_t = 1-\frac{t}{\mathrm{max} (1,\mathrm{steps} -1)} ,\mathrm{steps}  > 0$, $t$ is the current iteration step, and $\mathrm{steps}$ is the total iterations.
Based on this, the model focuses on unimodal features in the early iterations and emphasizes cross-modal features in the later stages.

Subsequently, the dominant and auxiliary modality features are determined based on the modality importance $\mu$, and $dom,aux_1,aux_2 \in \{\textbf{\textit{V}},\textbf{\textit{A}},\textbf{\textit{L}}\}$.
In the high-dimensional feature space, we utilize the phase difference between the dominant and auxiliary modalities to guide the auxiliary one in learning complementary information from the dominant modality.
To achieve this, we design a Decomposer module that adaptively models how the auxiliary modality aligns with the dominant one.
Specifically, at each iteration step $t$,  the dominant feature $f_{dom}^{fuse,t}$ and the auxiliary feature $f_a^{fuse,t}$ are concatenated to form a joint representation $[f_{dom}^{fuse,t}; f_{aux}^{fuse,t}] \in \mathbb{R}^{2D}$.
This representation is fed into a gating network, which predicts a projection weight $g_{aux}^t$ indicating the degree of phase alignment between the two modalities:
\vspace{-0.2em}
\begin{equation}
h_1 = \mathrm{ReLU}\big(W_1 [f_{dom}^{fuse,t}; f_{aux}^{fuse,t}] + b_1\big) \in \mathbb{R}^{D}, 
\quad
g_{aux}^t = \sigma(W_2 h_1 + b_2) \in \mathbb{R}^{D}
\end{equation}
where $\sigma(\cdot)$ denotes the Sigmoid activation function.
Accordingly, the dominant modality feature is projected into the auxiliary space:
\vspace{-0.15em}
\begin{equation}
    proj_{aux}^t = g_{aux}^t \odot f_{dom}^{fuse,t}
\end{equation}
The residual component of the auxiliary modality, representing information not captured by the projection:
\vspace{-0.55em}
\begin{equation}
    res_{aux}^t = f_{aux}^{fuse,t}-proj_{aux}^t
\end{equation}

\noindent
To balance alignment and complementarity, we impose a phase constraint on the orthogonality between the projection and residual terms:

\begin{equation}
    \mathcal{L} _{phase}^t = \frac{1}{N} \sum_{n \in aux}^{}\mathbb{E} [((proj_n^t)^\mathbf{T}res_n^t )^2] 
\end{equation}
Minimizing this loss encourages a moderate phase convergence, reducing excessive misalignment while preserving inter-modal complementarity.

However, the residual component may still contain noise.
To suppress such noise, a denoising network is applied to estimate the noise within the residual:
\vspace{-0.15em}
\begin{equation}
    noise_{aux}^t = \mathrm{Dropout} (\mathrm{ReLU} (W_{aux}res_{aux}^t))
\end{equation}
Finally, the cleaned residual and the projection are fused to form the refined auxiliary feature for the next iteration:
\vspace{-0.1em}
\begin{equation}
f_{aux}^{t+1} = proj_{aux}^t + \gamma (res_{aux}^t - noise_{aux}^t), 
\quad
f_{dom}^{t+1} = f_{dom}^{fuse,t}
\end{equation}
Here, $\gamma$ serves as a balance coefficient that controls the contribution of the denoised residual information to the auxiliary feature update.
In this way, the Decomposer enables each auxiliary modality to dynamically align with the dominant one, learn complementary information, and iteratively refine its representation under noise-suppressed guidance.

\vspace{-0.5em}
\subsection{Objective optimization}
\vspace{-0.2em}
After the final iteration, the refined representations from all three modalities $f_{dom}^{final}, f_{aux_1}^{final},f_{aux_2}^{final}$ are integrated to perform the final sentiment prediction:
\vspace{-0.6em}
\begin{equation}
    f_{final} = [f_{dom}^{final}, f_{aux_1}^{final},f_{aux_2}^{final}]
\end{equation}
The sentiment analysis objective is optimized using the cross-entropy loss:
\vspace{-0.6em}
\begin{equation}
    \mathcal{L}_{task} = - \frac{1}{N} \sum_{i=1}^{N} \sum_{c=1}^{C} y_{i,c} \log \left( \hat{y}_{i,c} \right),
\end{equation}
where $\hat{y}_{i,c} = \text{Softmax}(W f_{final}^{(i)} + b)_c$ denotes the predicted probability.
We integrate the above losses:
\vspace{-0.6em}
\begin{equation}
    \mathcal{L}_{total} = \mathcal{L}_{task} + \eta_1 \mathcal{L}_{uni} + \eta_2\mathcal{L}_{phase}
\end{equation}

\section{Experiment}
\vspace{-0.6em} 
\subsection{Datasets and Evaluation Metrics}
Datasets. We evaluate PRLF and representative MSA methods on three benchmark datasets: CMU-MOSI \cite{r44}, CMU-MOSEI \cite{r45}, and SIMS \cite{r56}.
CMU-MOSI includes 2,199 video clips with acoustic and visual features.
CMU-MOSEI consists of 22,856 YouTube review clips with features sampled at 20 Hz and 15 Hz.
SIMS is a Chinese multimodal sentiment dataset with 2,281 film and TV segments, providing aligned text, audio, and visual annotations.

\begin{table*}[t]
\vspace{-1.2em}
\centering
\caption{Performance comparison on CMU-MOSI and CMU-MOSEI datasets under different missing-modality conditions.}
\vspace{-0.3em}
\renewcommand\arraystretch{0.6}
\setlength{\tabcolsep}{1.1mm}
\small
\begin{tabular}{l lcccccccc}
\toprule
\textbf{Dataset} & \textbf{Models} & \textbf{\{l\}} & \textbf{\{a\}} & \textbf{\{v\}} & \textbf{\{l,a\}} & \textbf{\{l,v\}} & \textbf{\{a,v\}} & \textbf{Avg.} & \textbf{\{l,a,v\}} \\
\midrule
\multirow{8}{*}{\textbf{MOSI}} 
& Self-MM~\cite{r9} & 67.80 & 40.95 & 38.52 & 69.81 & 74.97 & 47.12 & 56.53 & 84.64 \\
& MISA~\cite{r4} &81.17&43.51&49.24&80.99&81.49&49.42&64.3&83.36\\
& TETFN~\cite{r55} &81.04&39.32&39.32&81.09&80.99&39.32&60.18& 82.33 \\ 
& MMIM~\cite{r54} &81.17&36.52&31.53&81.78&81.02&37.63&58.28 &82.28\\
& UMDF~\cite{r17} & 82.92 &67.80&59.92&\textbf{85.63}&84.09&72.98&75.56&83.36 \\
& HRLF~\cite{r19} &  83.36 &69.47 &\textbf{64.59 }&83.82 &83.56& 75.62 &76.74 &84.15 \\
& CorrKD \cite{r17} &81.20 & 66.52 & 60.72 & 83.56 & 82.41 & 73.74 & 74.69 & 83.94 \\
& EMOE~\cite{r46} &78.66 &65.64&58.1&83.52&80.24&73.69&73.31&85.4 \\
\rowcolor{blue!10}
& PRLF &\textbf{83.82} &\textbf{69.63}&64.05&84.98&\textbf{84.13}&\textbf{76.03}&\textbf{77.02}&\textbf{85.78} \\
\midrule
\multirow{8}{*}{\textbf{MOSEI}} 
& Self-MM~\cite{r9} & 71.53 & 43.57 & 37.61 & 75.91 & 74.62 & 49.52 & 58.79 & 83.69 \\
& MISA~\cite{r4} &80.55 & 58.99 & 58.99 &83.76 &83.46 &58.99 &70.79 &84.55 \\
&TETFN~\cite{r55} &81.88&58.99&58.98&83.86&83.55&58.98&71.04&84.83 \\  
& MMIM~\cite{r54} &80.96&59&59.28&81.09&81.29&59.43&70.18& 82.06 \\
& UMDF~\cite{r17} &81.57 &67.42&61.57&83.25&82.14&69.48&74.24&82.16 \\
& HRLF~\cite{r19} &  82.05& 69.32& \textbf{64.90} &82.62& 81.09 &73.80 &75.63& 82.93 \\
& CorrKD \cite{r17} & 80.76 & 66.09 & 62.30 & 81.74 & 81.28 & 71.92 & 74.02 & 82.16 \\
& EMOE~\cite{r46} &78.46 &65.33&61.54&81.88&81.26&70.57&73.17&85.3 \\
\rowcolor{blue!10}
& PRLF &\textbf{82.31} &\textbf{69.49}&63.67&\textbf{84.06}&\textbf{83.63}&\textbf{74.32}&\textbf{76.24}&\textbf{85.44 }\\
\bottomrule
\end{tabular}
\label{tab:missing_data_comparison}
\vspace{-1.0em}
\end{table*}

\textbf{Evaluation Metric.} We evaluate PRLF using the F1 score on three datasets, and additionally report its ACC-7, ACC-5, ACC-2, and MAE metrics on the same datasets in the supplementary material.

\textbf{Implementation details.}
For the CMU-MOSI and CMU-MOSEI datasets, we adopt 300-dimensional GloVe embeddings \cite{r48} and 768-dimensional BERT-base-uncased hidden states \cite{r49} for the language modality.
Visual features are obtained from Facet \cite{r50}, which provides 35-dimensional facial action unit representations, and acoustic features are extracted using COVAREP \cite{r51}, yielding 74-dimensional descriptors.
the detailed hyper-parameter settings are as follows: the balance coefficient $\gamma$ is 0.8, the loss weight $\eta_1, \eta_2$ are 0.5,0.1, the iteration step is 4.
The missing modality features are substituted with zero vectors.
During training, the seed changes with each epoch, resulting in different missing patterns for each sample per round, with all three modalities missing simultaneously and the missing rate kept constant. 
During testing, we follow the CorrKD \cite{r17} setup, and all results are averaged over five random seeds.  

\vspace{-0.6em}
\subsection{Comparison with State-of-the-art Methods.}
\textbf{Robustness to Inter-modality Missingness.}
As shown in Table~\ref{tab:missing_data_comparison}, PRLF consistently surpasses existing multimodal fusion methods on both CMU-MOSI and CMU-MOSEI datasets across all modality settings.
On MOSI, PRLF achieves the best overall accuracy, outperforming HRLF (76.74\%) and UMDF (75.56\%), while maintaining competitive results under single- and bimodal conditions, demonstrating strong modality-specific representation learning and robust cross-modal interaction.
On MOSEI, PRLF further achieves the highest average accuracy (76.24\%) and the best full-modality performance (85.44\%), confirming its superior adaptability to both complete and missing modality scenarios.
Similarly, as shown in Table~\ref{tab:ablation_sims}, PRLF also achieves the best or comparable performance under inter-modality missing conditions on the SIMS dataset, with the highest average accuracy of 81.19\%.
Compared with strong baselines such as LNLN (80.86\%) and EMOE (80.12\%), PRLF shows consistent improvements, particularly in settings (\{l,a\}, \{l,v\}, \{a,v\}).
These results demonstrate that the PRLF framework enhances cross-modal alignment, improves inter-modal complementarity, and maintains strong robustness under missing modality conditions.
\begin{table}[t]
\vspace{-0.8em}
\centering
\setlength{\tabcolsep}{5.2pt}
\renewcommand{\arraystretch}{0.88}
\caption{Performance comparison on SIMS under different missing conditions}
\vspace{-0.3em}
\label{tab:ablation_sims}
\begin{tabular}{lccccccc}
\hline
Models & \multicolumn{7}{c}{Testing Conditions}  \\ 
       & \{l\} & \{a\} & \{v\} & \{l,a\} & \{l,v\} & \{a,v\}  & Avg.\\
\hline
MISA \cite{r4} &75.15&56.95&56.82&77.07&75.17&56.78&66.32\\
Self-MM \cite{r9} &67.11&70.36&72.85&73.71&76.76&78.00&73.13\\
MMIM \cite{r54} &73.18&57.70&57.86&73.43&72.96&57.76&65.48\\
TETFN \cite{r55} &79.19&56.82&56.82&79.19&79.16&56.82&68\\
ALMT \cite{r53} &76.17&81.08&\textbf{81.91}&75.75&76.36&81.91&78.86\\
LNLN \cite{r15} &79.72&\textbf{81.91}&\textbf{81.91}&80.11&79.69&81.91&80.86\\
EMOE \cite{r46} &78.66&81.25&80.69&79.54&79.36&81.27&80.12\\
\rowcolor{blue!10}
PRLF &\textbf{80.28}&81.74&81.63&\textbf{80.57}&\textbf{80.36}&\textbf{82.58}&\textbf{81.19}\\
\hline
\end{tabular}
\vspace{-1.2em}
\end{table}

\begin{figure}[htbp]
\vspace{-1.29em}
  \centering
  \begin{subfigure}{1\linewidth}
    \includegraphics[width=\linewidth]{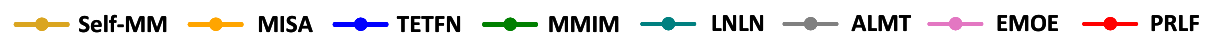}
    \label{fig:short-a}
  \end{subfigure}

  \vspace{-13pt}

  \begin{subfigure}{0.33\linewidth}
    \includegraphics[width=\linewidth]{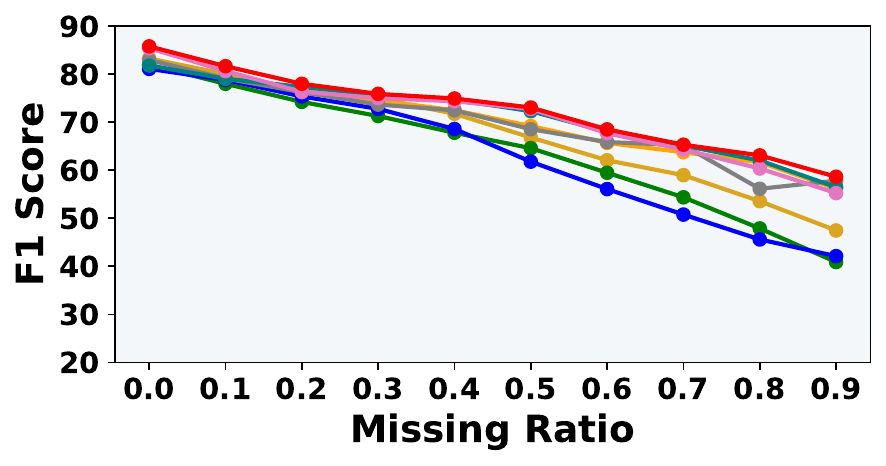}
    \caption{MOSI.}
    \label{fig:short-b}
  \end{subfigure}
  \hfill
  \begin{subfigure}{0.32\linewidth}
    \includegraphics[width=\linewidth]{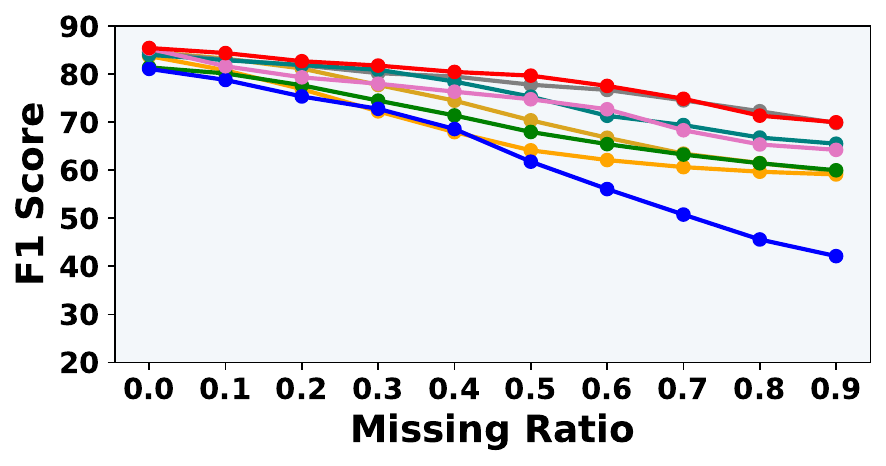}
    \caption{MOSEI.}
    \label{fig:short-c}
  \end{subfigure}
  \hfill
  \begin{subfigure}{0.32\linewidth}
    \includegraphics[width=\linewidth]{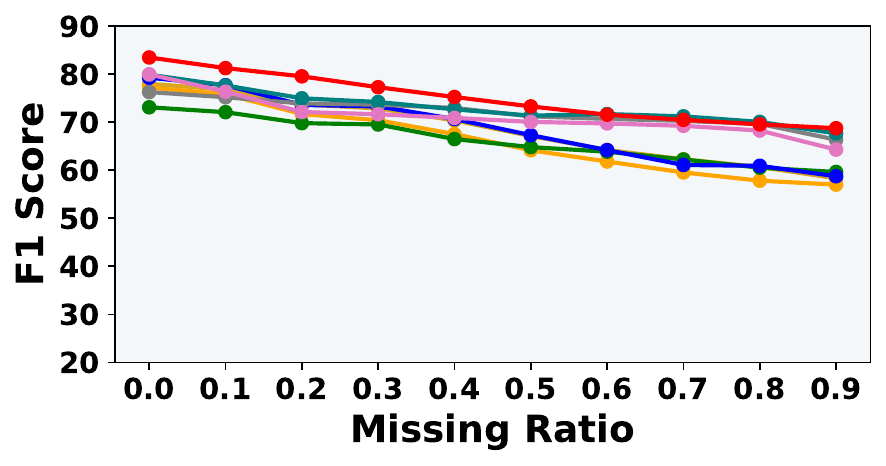}
    \caption{SIMS.}
    \label{fig:short-c}
  \end{subfigure}
  \vspace{-0.6em}
  \caption{Comparison results of intra-modality missingness. We report the F1 score.}
  \label{fig:short5}
  \vspace{-1.6em}
\end{figure}

\textbf{Robustness to Intra-modality Missingness.}
In this subsection, we compare PRLF with several representative methods. We randomly drop modal features at ratios of $p \in \{0,0.1,0.2,...,0.9\}$ to simulate intra-modal missingness in Fig.~\ref{fig:short5}. PRLF consistently outperforms other methods on all three datasets, with particularly significant advantages under high missing ratios, demonstrating stronger robustness to intra-modal missingness. Although the F1 scores of all methods decrease as missingness increases, PRLF shows the slowest performance degradation, indicating its superior ability to exploit residual information. When the missing ratio reaches 0.9, PRLF still achieves F1 scores of 60 on MOSI and 70 on MOSEI, both substantially higher than TETFN, highlighting its robustness under extreme conditions.
The detailed results are reported in supplementary Tables 9 and 10, and Table 11. Fusion-oriented methods such as EMOE perform well at low missing ratios but degrade rapidly as missingness increases, while incomplete-data methods like LNLN are more robust under severe missingness yet still inferior to PRLF. By dynamically estimating modality reliability via FIM and CC, PRLF maintains stable performance across all missing rates.
On MOSI, MOSEI, and SIMS, the averaged F1 scores of PRLF, EMOE, and LNLN are (\textbf{72.48}, 71.21, 71.25), (\textbf{78.82}, 74.60, 75.65), and (\textbf{75.04}, 71.25, 73.13), respectively, further confirming PRLF’s consistent overall advantage across varying levels of intra-modal missingness.

\begin{table*}[htbp]
\vspace{-1.3em}
\centering
\caption{Iteration Step Ablation results for the intra-modality missingness on MOSI. }
\label{tab:ablation_mosi}
\renewcommand\arraystretch{0.84}
\setlength{\tabcolsep}{1.2mm}
\begin{tabular}{lcccccccccc}
\hline
\multirow{2}{*}{Iteration} & \multicolumn{10}{c}{Missing rate $p$}  \\
      &$0 $& $0.1$ & $0.2$ & $0.3$ & $0.4$ & $0.5$ & $0.6$  & $0.7$ & $0.8$ & $0.9$\\
       \hline
$Step=1$ &81.14&79.69&75.44&72.25&71.09&69.58&64.71&60.73&59.91&55.37\\
$Step=2$ &83.08&80.39&76.87&73.88&72.34&70.65&65.82&62.06&61.15&56.39 \\
$Step=3$ &85.21&81.29&\textbf{78.06}&75.28&73.96&71.97&67.75&64.62&62.23&57.59 \\
\rowcolor{blue!10}
$Step=4$ &\textbf{85.78}&\textbf{81.65}&77.99&\textbf{75.89}&\textbf{74.93}&\textbf{73.05}&\textbf{68.5}&\textbf{65.31}&\textbf{63.09}&\textbf{58.64} \\
$Step=5$ &84.49&81.08&77.29&75.19&73.78&71.24&67.27&64.16&62.33&57.48 \\
\hline
\end{tabular}
\label{tab:3}
\vspace{-2.99em}
\end{table*}

\begin{figure}[htbp]
\vspace{-1em}
  \centering
  \begin{subfigure}{0.49\linewidth}
    \includegraphics[width=\linewidth]{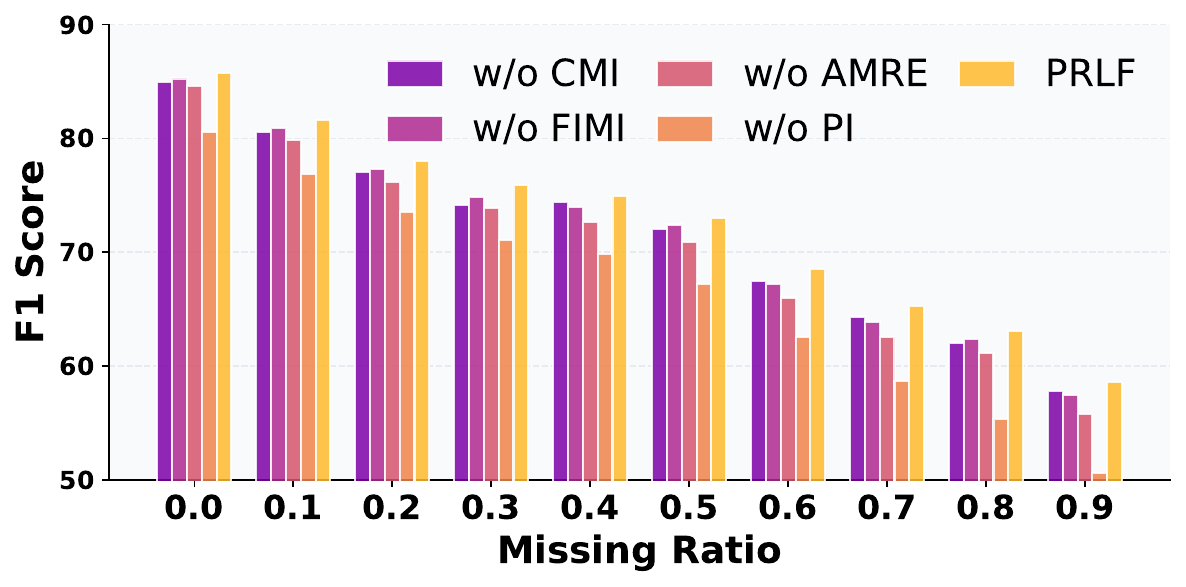}
    \caption{Module Ablation.}
    \label{fig:short6-a}
  \end{subfigure}
  \hfill
  \begin{subfigure}{0.49\linewidth}
    \includegraphics[width=\linewidth]{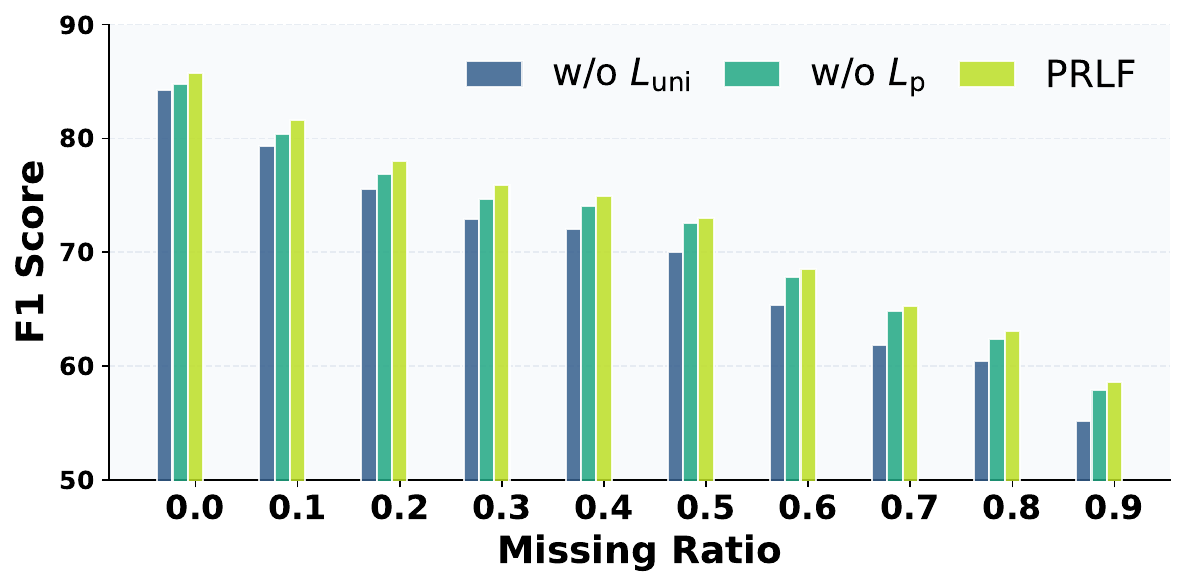}
    \caption{Loss Ablation.}
    \label{fig:short6-b}
  \end{subfigure}
  \caption{Ablation for the intra-modality missingness on MOSI. }
  \label{fig:short6}
  \vspace{-2.9em}
\end{figure}

\subsection{Ablation Studies}
\vspace{-0.3em}
\textbf{Iteration Step Ablation on Inter- and Intra-modality Missingness.}
As shown in Table~\ref{tab:3}, ~\ref{tab4},
the experiments on both inter-modal and intra-modal missing scenarios exhibit a consistent trend: as the number of iterations increases from 1 to 4, the model performance steadily improves, reaching its peak at Step = 4. 
This indicates that multi-step progressive interaction effectively enhances feature alignment and strengthens the representations of incomplete or missing modalities by leveraging guidance from the dominant modality, particularly under high missing rates or partial modality inputs. 
However, when the number of iterations increases to 5, performance begins to decline, suggesting that excessive interaction steps may impair the model’s generalization ability. 
Therefore, Step = 4 achieves the optimal balance between representational power and stability, confirming the effectiveness and rationality of the proposed progressive fusion strategy for handling diverse missing-modality conditions.

\begin{table}[htbp]
\vspace{-1.0em}
\centering
\setlength{\tabcolsep}{4.5pt}        
\renewcommand{\arraystretch}{0.85}   
\caption{Iteration Step Ablation results for the inter-modality missingness on MOSI.}
\vspace{-0.6em}                      
\label{tab:ablation_mosi}
\footnotesize                        
\begin{tabular}{lcccccccc}
\hline
\multirow{2}{*}{\makecell[c]{Iteration}} & \multicolumn{7}{c}{Testing Conditions}  \\ 
       & \{l\} & \{a\} & \{v\} & \{l,a\} & \{l,v\} & \{a,v\} & Avg. & \{l,a,v\}\\
\hline
$Step=1$ &79.62&64.33&59.68&80.47&80.06&69.57&72.29&81.14 \\
$Step=2$ &81.49&66.87&62.34&82.65&82.33&72.64&74.72&83.08 \\
$Step=3$ &82.91&68.73&63.11&84.36&\textbf{84.27}&75.58&76.49&85.21 \\
\rowcolor{blue!10}
$Step=4$ &\textbf{83.82} &\textbf{69.63}&\textbf{64.05}&\textbf{84.98}&84.13&\textbf{76.03}&\textbf{77.11}&\textbf{85.78} \\
$Step=5$ &82.47&68.51&62.77&83.54&82.61&74.88&75.80&84.49 \\
\hline
\end{tabular}
\label{tab4}
\end{table}

\begin{table}[htbp]
\vspace{-0.3em}
\centering
\setlength{\tabcolsep}{3.6pt}
\renewcommand{\arraystretch}{0.88}
\caption{Module Ablation for inter-modality missingness on MOSI. Confidence-based Modality Importance (CMI), Fisher Information-based Modality Importance (FIMI), Adaptive Modality Reliability Estimator (AMRE), Progressive Interaction (PI).}
\vspace{-0.7em}
\label{tab:ablation_mosi}
\footnotesize
\begin{tabular}{lcccccccc}
\hline
Models & \multicolumn{7}{c}{Testing Conditions}  \\ 
       & \{l\} & \{a\} & \{v\} & \{l,a\} & \{l,v\} & \{a,v\} & Avg. & \{l,a,v\}\\
\hline
\rowcolor{blue!10}
PRLF &\textbf{83.82} &\textbf{69.63}&\textbf{64.05}&\textbf{84.98}&\textbf{84.13}&\textbf{76.03}&\textbf{77.11}&\textbf{85.78} \\
w/o CMI &82.97&68.54&63.21&83.74&83.25&74.98&76.12&85.07 \\
w/o FIMI &83.04&68.77&63.25&83.41&82.99&75.06&76.09&84.88 \\
w/o AMRE &81.92&67.58&61.95&82.36&82.15&73.79&74.96&83.05 \\
w/o PI &78.48&63.53&58.76&79.66&79.01&68.49&71.32&79.33 \\
\hline
\end{tabular}
\vspace{-1.8em}
\label{table:short5}
\end{table}

\begin{table}[htbp]
\vspace{-1.8em}
\centering
\setlength{\tabcolsep}{3.5pt}
\renewcommand{\arraystretch}{0.88}
\caption{Loss ablation on MOSI (inter-modality missingness).}
\vspace{-0.7em}
\label{tab:ablation_mosi}
\footnotesize
\begin{tabular}{lcccccccc}
\hline
Models & \multicolumn{7}{c}{Testing Conditions}  \\ 
       & \{l\} & \{a\} & \{v\} & \{l,a\} & \{l,v\} & \{a,v\} & Avg. & \{l,a,v\}\\
\hline
\rowcolor{blue!10}
PRLF &\textbf{83.82} &\textbf{69.63}&\textbf{64.05}&\textbf{84.98}&\textbf{84.13}&\textbf{76.03}&\textbf{77.11}&\textbf{85.78} \\
w/o $\mathcal{L}_{uni}$ &82.64&68.32&62.88&83.07&82.98&74.51&75.73&84.81 \\
w/o $\mathcal{L}_{phase}$ &83.07&69.24&63.58&84.75&83.89&75.26&76.63&85.49\\
\hline
\end{tabular}
\label{tab:short6}
\vspace{-1.5em}
\end{table}

\textbf{Module Ablation on Inter- and Intra-modality Missingness.}
As shown in Table~\ref{table:short5} and Fig.~\ref{fig:short6-a}, experiments on both inter- and intra-modality missing scenarios consistently demonstrate that the CMI, FIMI, ANRE, and PI are indispensable components of the PRLF framework.
Removing the PI module leads to the most pronounced performance degradation across all settings, particularly under high intra-modality missing rates.
This highlights its crucial role in cross-modal alignment and fusion through iterative optimization and dominant-modality guidance.
Eliminating the AMRE module results in consistent performance drops, as the model loses the ability to dynamically identify and exploit the most reliable modality for fusion—underscoring the superiority of adaptive weighting over static strategies.
The CMI and FIMI serve as complementary mechanisms for assessing modality reliability, with FIMI showing slightly greater importance, especially under degraded input conditions.
Notably, the substantial performance gaps observed in bimodal and full-modality settings further confirm that PI and AMRE are essential for optimizing multimodal collaboration and enhancing robustness against both partial and complete modality failures.

\textbf{Loss Ablation on Inter- and Intra-modality Missingness.}
As shown in Table~\ref{tab:short6} and Fig.~\ref{fig:short6-b}, removing either the feature uniformity loss ($\mathcal{L}_{uni}$) or the phase consistency loss ($\mathcal{L}_{phase}$) results in consistent performance degradation under both inter- and intra-modality missingness.
$\mathcal{L}_{uni}$ encourages stable feature representations across different missing conditions, while $\mathcal{L}_{phase}$ preserves feature directions among modalities.
Their combination enables PRLF to better resist noise and retain discriminative structures, especially under high missing ratios.


\begin{figure}[htbp]
  \centering

  \begin{subfigure}{0.3\linewidth}
    \includegraphics[width=\linewidth]{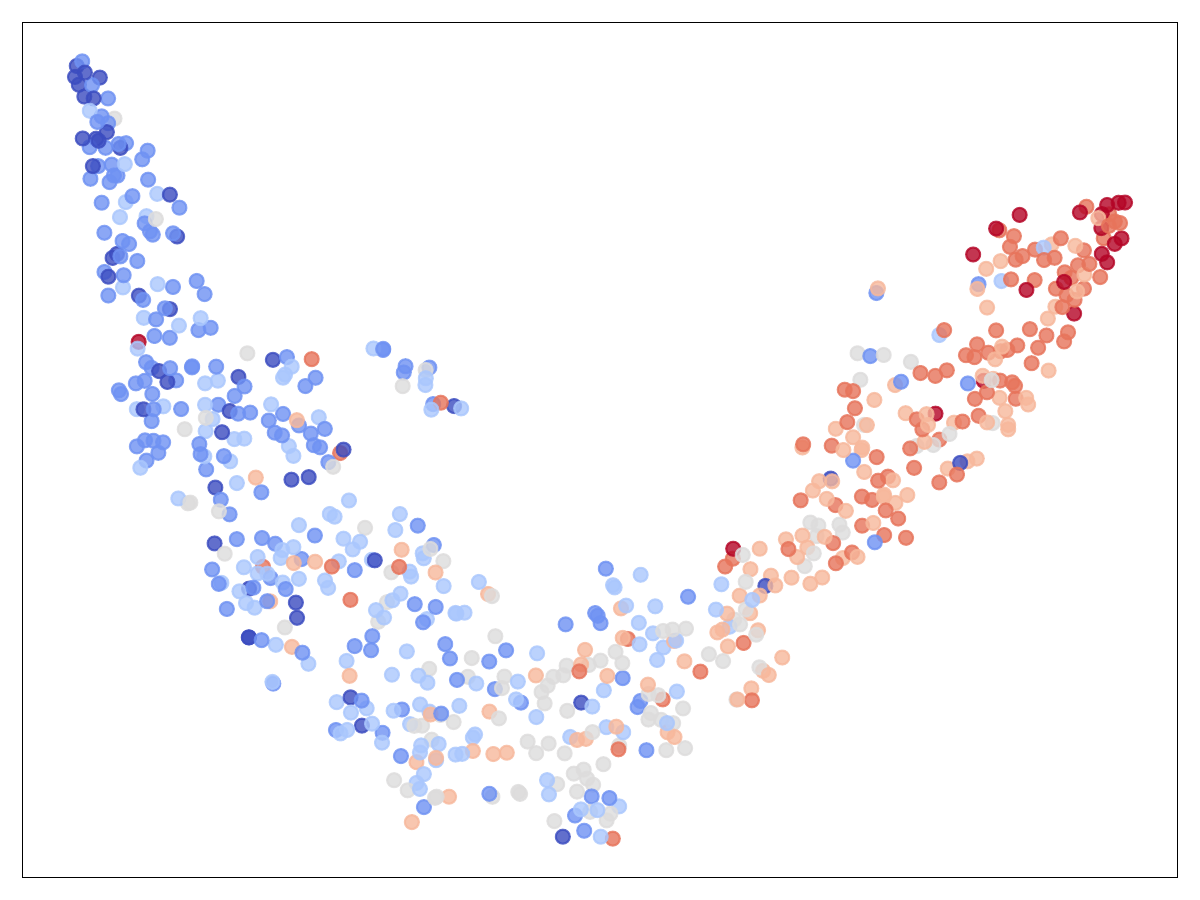}
    \caption{PRLF (w/o AMRE).}
    \label{fig:short-b}
  \end{subfigure}
  \hfill
  \begin{subfigure}{0.3\linewidth}
    \includegraphics[width=\linewidth]{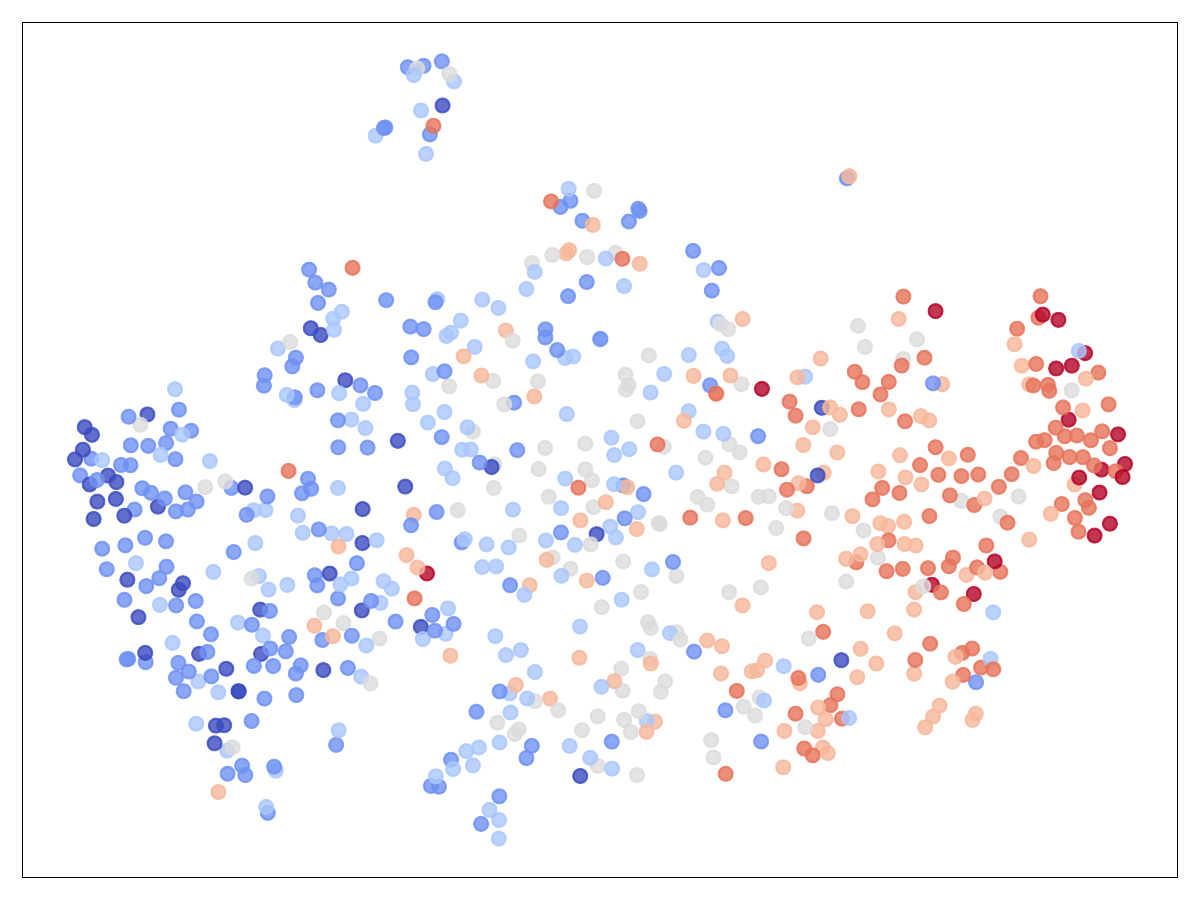}
    \caption{PRLF (w/o PI).}
    \label{fig:short-c}
  \end{subfigure}
  \hfill
  \begin{subfigure}{0.3\linewidth}
    \includegraphics[width=\linewidth]{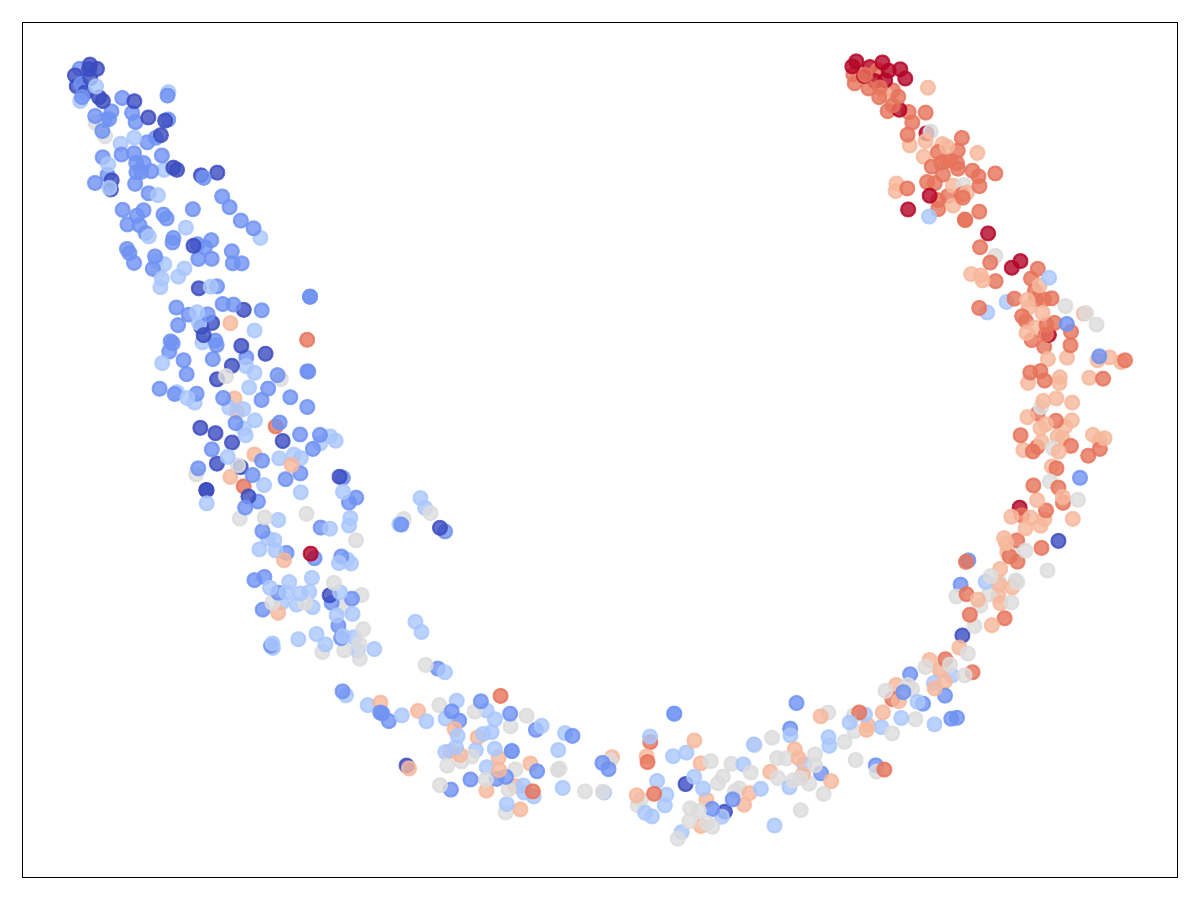}
    \caption{PRLF.}
    \label{fig:short-c}
  \end{subfigure}
\hfill
    \begin{subfigure}{1\linewidth}
    \includegraphics[width=\linewidth]{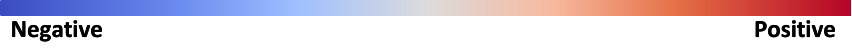}
    \label{fig:short-a}
  \end{subfigure}
\vspace{-3.2em}
  \caption{  T-SNE visualization on MOSI. Default: intra-modality missingness ($p=0.5$). }
  \label{fig:short7}
  \vspace{-0.5em}
\end{figure}

\begin{figure}
  \centering
  \begin{subfigure}{0.34\linewidth}
    \includegraphics[width=\linewidth]{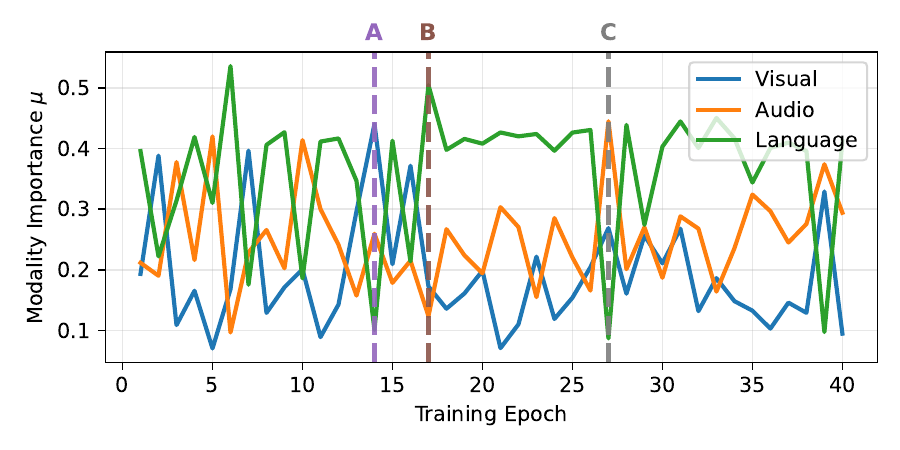}
  \end{subfigure}
  \begin{subfigure}{0.64\linewidth}
    \includegraphics[width=\linewidth]{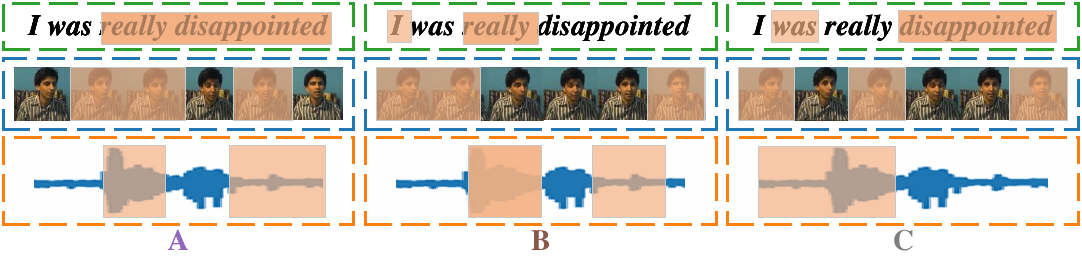}
  \end{subfigure}
  \vspace{-0.8em} 
  \caption{Modality importance $\mu$ changes (left) and different weights for the same sample under varying missing data (right).}
  \label{fig8:short}
  \vspace{-1.9em}
\end{figure}



\textbf{Visualization of Feature Distribution.} 
As illustrated in Fig.~\ref{fig:short7}, AMRE evaluates modality reliability and promotes more compact feature distributions. The clustering variances with and without AMRE are 1.205 and 1.367, respectively, indicating more pronounced convergence when AMRE is applied. Compared with the complete model, the ablated variants exhibit clear distribution discrepancies. Without AMRE, the clusters become noticeably less compact, reflecting the model’s inability to accurately estimate modality reliability and suppress unreliable features. Removing the PI module further results in blurred class boundaries, suggesting insufficient cross-modal alignment. In contrast, the full PRLF framework produces more compact and semantically consistent feature distributions that align well with emotion intensity. These results demonstrate that AMRE mitigates noise from unreliable modalities, while PI progressively enhances cross-modal semantic alignment under missing-modality conditions.

\textbf{Dynamic Modality Importance under Varying Missing Conditions.} Fig. ~\ref{fig8:short} visualizes the dynamic evolution of the modality importance weight $\mu$ and presents representative examples under different missing conditions. It can be observed that modality reliability varies significantly across different missing scenarios. The proposed method effectively evaluates the validity of each modality in an adaptive manner and accordingly selects the dominant modality to guide the subsequent progressive interaction and feature alignment process.
\vspace{-0.5em}

\section{Conclusion}
\vspace{-0.3em}
We presented PRLF, a progressive representation learning framework that addresses the challenges of multimodal sentiment analysis under uncertain missing-modality conditions.
By integrating an Adaptive Modality Reliability Estimator (AMRE) and a Progressive Interaction (ProgInteract) mechanism, PRLF dynamically evaluates modality reliability and progressively aligns cross-modal representations.
This design suppresses noise from incomplete modalities and enhances semantic consistency across views.
Experiments on multiple benchmarks demonstrate that PRLF delivers reliable emotion recognition and stable cross-modal fusion under various missing-modality conditions, advancing the robust Multimodal Sentiment Analysis.


%
%
\bibliographystyle{splncs04}
\bibliography{main}
\end{document}